%% file: root.tex
\documentclass[letterpaper, 10 pt, conference]{ieeeconf}  

\IEEEoverridecommandlockouts                              

\usepackage[noadjust]{cite}
\usepackage{algorithm2e}
\usepackage{amsmath,amssymb}
\usepackage{booktabs}
\usepackage{float}
\usepackage{cleveref} 

\usepackage{dblfloatfix}

\def\etal{\emph{et al.}}
\usepackage{flafter}

\makeatletter
\renewcommand{\paragraph}{
  \@startsection{paragraph}{4}%
  {\z@}{0.5em}{-1em}%
  {\normalfont\normalsize\bfseries}%
}
\makeatother

\usepackage{graphics} 
\usepackage{epsfig} 
\usepackage{amsmath} 
\usepackage{amssymb}  

\title{\LARGE \bf
Real Time Monocular Vehicle Velocity Estimation using Synthetic Data
}

\author{Robert McCraith$^{1}$, Lukas Neumann$^{1}$, Andrea Vedaldi$^{1}$
\thanks{Authors are members of the Visual Geometry Group in the University of Oxford, UK.
emails: {\tt\small robert, lukas, vedaldi @robots.ox.ac.uk}
}
}

\begin{document}

\maketitle
\thispagestyle{empty}
\pagestyle{empty}

\begin{abstract}
\input{abstract}
\end{abstract}
\input{intro}

\input{relatedwork}

\input{method}

\input{experiments}
\input{conclusion}

\bibliography{egbib}
\bibliographystyle{plain}

\end{document}

%% file: abstract.tex
Vision is one of the primary sensing modalities in autonomous driving.
In this paper we look at the problem of estimating the velocity of road vehicles from a camera mounted on a moving car.
Contrary to prior methods that train end-to-end deep networks that estimate the vehicles' velocity from the video pixels, we propose a two-step approach where first an off-the-shelf tracker is used to extract vehicle bounding boxes and then a small neural network is used to regress the vehicle velocity from the tracked bounding boxes.
Surprisingly, we find that this still achieves state-of-the-art estimation performance with the significant benefit of separating perception from dynamics estimation via a clean, interpretable and verifiable interface which allows us distill the statistics which are crucial for velocity estimation.
We show that the latter can be used to easily generate synthetic training data in the space of bounding boxes and use this to improve the performance of our method further.



%% file: intro.tex
\section{Introduction}\label{sec:intro}

Autonomous driving systems rely on a wide array of sensors including LiDARs, radars and cameras.
LiDAR sensors are especially good at estimating the position and velocities of vehicles and obstacles ($0.25$m/s at the distance of $15$ meters~\cite{kellner2013instantaneous}, $0.71$m/s at a wider range of distances~\cite{CVPR2017Challenge}).
However, LiDAR sensors are also very expensive, not reliable in adverse weather conditions~\cite{kutila2016automotive} and easily confused by exhaust fumes~\cite{hasirlioglu2017effects}.
Depending on a single source of information in a context such as autonomous driving is also inherently fragile.
Hence, it is natural to develop alternative sensors either as redundancies or to improve the accuracy and robustness of sensors like LiDARs.
Cameras are a natural choice as they are very cost effective and in principle sufficient for navigation given that humans drive cars primarily using their sense of vision.
Furthermore, there is a substantial amount of computer vision research that is directly applicable to autonomous driving.

 
In this paper, we aim to predict the velocity of other cars based only on a video from a standard monocular camera.
Because the vehicle where the camera is mounted on (the \emph{ego-vehicle}) is typically moving as well, the velocity estimate of other vehicles is relative to the velocity of the ego-vehicle.

Our main contribution is to show that, for this problem, one can separate perception from velocity estimation by first mapping the visual data to a mid-level representation --- the space of vehicle bounding boxes --- with no loss in performance compared to much more complex estimation approaches.
The velocity estimation problem is then modelled purely on bounding boxes, which change their position and size over time, tracking the motion of vehicles on the road in a simplified view of the data.
Using such an approach, our simple model outperforms the winning model~\cite{kampelmuhler2018} of CVPR 2017 Vehicle Velocity Estimation Challenge~\cite{CVPR2017Challenge}, which is a much more complex model that combines tracking with two different deep networks for monocular depth and optical flow estimation.

A major advantage of using such a simple intermediate representation is that it becomes much easier to \emph{simulate} training data for the velocity estimator model.
Still, we show that doing so in an effective manner requires to capture accurately the statistics of real bounding boxes.
Our second contribution is thus to show how such a synthetic dataset is created, by extracting the necessary data priors from a small set of real data and using it to generate a much larger training set in the mid-level feature space.
We then show that training only using this synthetic dataset results in excellent performance on real test data.
In the process, we also distill the data statistics that are crucial for velocity estimation from visual data, clarifying in the process what is important and what information is not for this task.

The ability to easily generate synthetic data is not only beneficial to improve velocity estimation accuracy at virtually no cost for existing scenarios, but can be also used to train the model for different driving scenarios, such as different countries, and to model different driving styles and various emergency situations (e.g.~unexpected breaking) without the need for laborious real data collection or for expensive 3D photo-realistic animation in order to cater for such situations.

Our approach can also be seen as encapsulating perception, which is often implemented by opaque and difficult-to-diagnose components such as deep convolutional neural networks, in a module which has a simple, interpretable, and testable interface.
Decomposing systems in modules that are individually verifiable and that can be connected in a predictable manner is essential for autonomous systems to meet reliability standards such as ISO 26262~\cite{ISO26262, palin2011iso}.
Hence, while end-to-end trainable systems may be conceptually preferable, decompositions such as the one we propose may be essential in practice --- fortunately, as we show, this may come with no loss of performance.

The rest of the paper is structured as follows. In \cref{sec:relatedwork}, prior work is discussed. In \cref{sec:method}, we introduce the method, in \cref{sec:dataset} we discuss the data generation. The evaluation is presented in \cref{sec:evaluation} and the paper is concluded in \cref{sec:conclusion}.

%% file: relatedwork.tex
\section{Related Work}\label{sec:relatedwork}
\input{fig-samples.tex}
\paragraph{Depth and ego-motion estimation}

In recent years numerous approaches to monocular depth and ego-motion estimation have been explored thanks to the expressive power of convolutional neural networks. Supervised approaches~\cite{eigen2014depth, ummenhofer2017demon, xu2017multi} depend on pixel-wise depth annotations to be available for each pixel of the training set, which results in costly data collection. Unsupervised methods~\cite{zhou2017unsupervised, Klodt18} on the other hand depend heavily on camera intrinsics and rely on camera motion between successive frames to produce coarse relative depths, which are difficult to use for predicting exact distances. 

\paragraph{Velocity estimation}

Classical vision techniques for motion detection from a moving camera such as Yamaguchi \etal \cite{yamaguchi2006} first match points in successive frames and then filter them based on their location and compatibility with the epipolar geometry.
A similar approach is used in Fanani~\etal\cite{IMOEpipolarCNN}, where candidate objects are first filtered by a CNN which detects vehicles and then tests based on the epipolar geometry are used to determine whether these vehicles are moving or not.

The above approaches aim to only differentiate between static and moving vehicles with no indication of their velocity.
Kampelm\"uhler \etal\cite{kampelmuhler2018}, the winner of CVPR 2017 Vehicle Velocity Estimation Challenge~\cite{CVPR2017Challenge}, extends these approaches to predict the relative velocity of the vehicles in view of the ego-vehicle.
This is achieved by passing the sequence of images through a depth~\cite{monodepth17} and flow~\cite{FlowNet} estimation network, as well as applying a classical tracking system~\cite{MedianFlow}.
The features extracted from these three sub-components are then concatenated and fed to a small neural network which predicts the velocity (see \cref{f:model} top).

\paragraph{Tracking}

Object tracking is a classical computer vision problem~\cite{lucas1981iterative}.
The multiple instance learning tracker~\cite{babenko2009visual} expresses the problem as a classification task, where detections in previous frames are used as training data for future frames.
MedianFlow~\cite{MedianFlow} uses the forward-backward error to validate which points are robust predictors of object movement.
The method is further improved in the TLD tracker~\cite{kalal2012tracking} by disabling online learning when the object is occluded and by allowing the algorithm to re-detect the object once it appears again.
More recently, with the emergence of deep learning, specialized networks have been trained to explicitly track location of objects in video sequences~\cite{held2016learning}.
Similarly, Siamese CNNs~\cite{bertinetto2016fully,valmadre2017end} have been exploited to build a powerful embedding to discriminate whether an image patch contains the same object or not, therefore tracking objects by appearance similarity.
For a systematic evaluation of tracking algorithms, we refer the reader to the VOT challenge results~\cite{kristan2017visual}.

\paragraph{Synthetic training data}

Synthetic training data have been successfully applied in various domains of computer vision, ranging from scene text detection~\cite{gupta2016synthetic} to optical flow estimation~\cite{FlowNet}.
In the autonomous driving domain, the most widely used synthetic dataset is Virtual~KITTI~\cite{Gaidon:Virtual:CVPR2016}, which contains 50 photo-realistic videos generated by the Unity game engine.
Thanks to the synthetic source of the data, the dataset comes with pixel-level annotations for segmentation, optical flow and depth, which would be virtually impossible and very expensive to achieve in real data.
More recently, the SYNTHIA~\cite{Ros_2016_CVPR} and Synscapes~\cite{wrenninge2018synscapes} artificial driving datasets have also been introduced.

The main difference to our work is that the above datasets are based on photo-realistic representation of the world, whereas our mid-level representation consists of mere bounding boxes, which are significantly easier to generate.

%% file: fig-samples.tex
\begin{figure*}
\centering
\begin{tabular}{cc}
\includegraphics[width=0.47\textwidth]{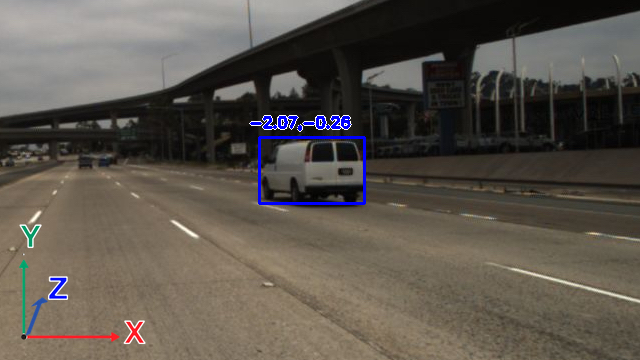} &
\includegraphics[width=0.47\textwidth]{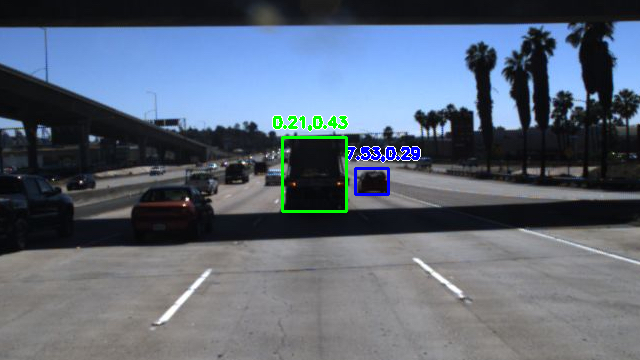}
\end{tabular}
\caption{Sample sequences from the TuSimple dataset at the last frame with their respective bounding box and velocity annotation in $Z$ and $X$ direction.}\label{f:samples}
\end{figure*}

%% file: method.tex
\section{Method}\label{sec:method}

\input{fig-model}

Our goal is to estimate the velocity of other vehicles imaged from a camera rigidly mounted on the ego-vehicle and looking forward.
We can model this situation as follows.
The input to the model is a sequence $\mathbf{I}=(I_1,\dots,I_T)$ of $T$ video frames extracted from the camera.
We also assume to have a bounding box $b_T$ tightly enclosing the vehicle of interest at the end of the sequence $T$.
The output of the model $\Phi$ is a 2D vector $V = \Phi(\mathbf{I},b_T)\in\mathbb{R}^2$ representing the velocity of the target vehicle at time $T$ projected on the ground plane relative to the ego-vehicle.\footnote{Estimating the vertical velocity $Y$ is essentially irrelevant for this application.}

\subsection{Geometry and elementary velocity estimation}\label{sec:elementary}

Next, we describe in some detail the geometry of the problem and provide a na\"ive solution based solely on projecting bounding boxes into 3D world coordinate system.
The physical constraint of the setup results in several simplifications compared to the general imaging scenario.
Let $P = (X,Y,Z) \in\mathbb{R}^3$ be a 3D point expressed in the ego-vehicle reference frame.
We can assume that the camera is at a fixed height $H$ from the ground looking straight ahead.
Hence, point $P$ projects to the image point
$
p = (u,v),
$
$
u = f\,\frac{X}{Z},
$
$
v = f\,\frac{Y+H}{Z}
$
where $f$ is the focal length of the camera.\footnote{The image coordinates $(u,v)$ are standardized, with $(0,0)$ corresponding to the view direction, and are in practice related to pixels coordinates via a non-linear transformation that accounts for the camera intrinsic parameters, including effects such as radial distortion.
We assume that the intrinsic parameters are known and that their effects has already been removed from the data.}

Now assume that the bounding box $b=(x,y,w,h) \in\mathbb{R}^4$ is given by the image coordinates $(u,v) = (x+w/2, y+h)$ of the mid point of the bottom edge of the box and by the box width and height $(w,h)$.
To a first approximation, we can assume that $(u,v)$ is the image of a certain virtual 3D point $P=(X,0,Z)$ rigidly attached to the vehicle at ground level.
Hence, since we know the height $H$ of the camera, we can readily infer the depth or distance $Z$ of the vehicle, and hence the 3D point, as
\begin{equation}\label{e:naive}
 Z = fH\,\frac{1}{v},
 \quad
 X = H\,\frac{u}{v}.
\end{equation}
If we can track the bounding box $b_t$ over time $t\in[1,T]$, then we can use~\cref{e:naive} to estimate the coordinates $P_t=(X_t,0,Z_t)$ of the 3D point relative to the ego-vehicle and obtain the velocity as the derivative $V = (\dot X_T, \dot Z_T)$.
However, owing to the varying quality of road surfaces and inclines or declines along a road this technique provides very poor estimate of vehicle velocity, especially for vehicles which are further away (see \cref{tbl:results}).
Indeed, analyzing \cref{e:naive} and assuming a camera with standard image resolution, we come to a conclusion that for vehicles in distance $d > 20$m the approach requires sub-pixel accuracy of the bounding-box estimate for a reasonable estimate of 3D position and hence velocity, which simply is not realistic.

\input{fig-bb-distribution}

\subsection{Deep learning for velocity estimation}%
\label{sec:velocityestimation}

Sophisticated models which combine several streams of information  from image pixels (depth, optical flow) perform significantly better, so one may be tempted to ascribe the poor performance of the geometric approach to the fact that too much information is discarded by looking at bounding boxes only.
We show that, somewhat surprising, \emph{this is not the case}.
Instead, we show below that bounding boxes \emph{are} sufficient provided that the modelling of dynamics is less na\"\i{}ve.

\paragraph{State-of-the art baseline}

Our reference model is the one of Kampelm\"{u}hler~\etal~\cite{kampelmuhler2018}.
They propose a complex network that combines three data streams, all derived from the video sequence $\mathbf{I}$ (see~\cref{f:model} top).
One stream applies a pre-trained monocular depth-estimation network, called MonoDepth~\cite{monodepth17}, to estimate a 3D depth map from the video.
Another stream applies FlowNet2~\cite{FlowNet}, a state-of-the-art optical flow estimation network, to estimate instead optical flow.
The last stream uses an off-the-shelf tracker~\cite{MedianFlow} to track the bounding box $b_T$ backward through time, and thus obtain a simplified representation $(b_1,\dots,b_T)$ of the vehicle trajectory in the image.
The output of the different streams are concatenated and passed through a multi-layer perceptron (MLP) to produce the final velocity estimate $V = \Phi(\mathbf{I},b_T)$.
In addition to pre-training MonoDepth, FlowNet2, and the tracker on external data-sources, the MLP, fusing the information, is trained on an ad-hoc benchmark dataset that contains vehicle bounding boxes for one frame with their ground-truth velocities and positions measured via a LiDAR (\cref{sec:dataset}).

Furthermore, three individual models of different-sized MLPs are used to determine velocity within the different vehicle distance ranges used in the evaluation. These range from 3 layers of 40 hidden neurons to 4 layers of 70 hidden neurons with increasing model size for vehicles further away~\cite{kampelmuhler2018}. The final velocity prediction is given by averaging output of 5 model instances, therefore in total 15 distance-specific models are used at testing time.

\paragraph{Our Model}
Similar to \cite{kampelmuhler2018} we use a fully connected neural network with 4 hidden layers of 70 neurons each with CReLU\cite{CReLU} activation and Dropout\cite{dropout} between layers during training. This allows us to reduce the runtime of our method significantly as we do not require estimation of optical flow and depth which allows our runtime to reduce from $423ms$ to $10ms$ and only one model is trained reducing the inference time further. 

\paragraph{Estimating velocity from bounding boxes}

Next, we describe our architecture to estimate vehicle velocity (see \cref{f:model} bottom).
We consider an input video sequence $\mathbf{I}=(I_1,\dots,I_T)$ of $T$ video frames capturing a vehicle at time $t\in[1,T]$.
As shown above, in the reference model, the velocity estimate $V$ is then obtained as
\begin{equation}\label{e:phi}
V = \Phi(\mathbf{I}, b_T)
\end{equation}
where $\Phi$ is neural network or a combination of several neural networks as in \cite{kampelmuhler2018}, taking video sequence $\mathbf{I}$ as an input and a bounding box $b_T$ in the last frame.

In our paper, we decompose $\Phi$ into two mappings $\Psi$ and $\Xi$ 
\begin{equation}\label{e:psi}
    V = \Psi(\Xi(\mathbf{I},b_T))    
\end{equation}

by introducing an intermediate representation $\mathbf{b}=(b_1,\dots,b_T)$ representing the vehicle image location at time $t\in[1,T]$
$
\mathbf{b} = \Xi(\mathbf{I},b_T), V = \Psi(\mathbf{b}),
$
where $\Xi$ is a off-the-shelf tracker component~\cite{MedianFlow} and $\Psi$ is the vehicle velocity estimator we train.

Using the above decomposition, we only require $(\mathbf{b},V)$ as supervision pairs, and not the $(\mathbf{I}, V)$ pairs as in the original formulation~\cite{kampelmuhler2018}, which are extremely expensive to obtain as it implies capturing and annotating many driving video sequences.

As before, bounding boxes $\mathbf{b}$ are represented as quadruples $(x,y,w,h)$.
The trajectories $\mathbf{b}$ are passed to a filter $g(\mathbf{b})$ that applies temporal Gaussian smoothing to each coordinate independently.
Finally, the output of the filter is flattened to a $4T$-dimensional vector and fed to a multi-layer perceptron $\bar \Psi : \mathbb{R}^{4T} \rightarrow \mathbb{R}$.
Hence, the overall model at inference time can be written as
$
 V 
 = \Psi(\mathbf{b}) 
 = \bar \Psi(\operatorname{vec}(g(\Xi(\mathbf{I},b_T)))).
$

At training time, assuming the tracker $\Xi$ is fixed, we however only need to train $\Psi$, using the pairs $(\mathbf{b},V)$ by minimizing the loss $\|V - \Psi(\mathbf{b})\|^2$. Next we show how the pairs can be obtained.

%% file: fig-model.tex
\begin{figure*}
\centering
\begin{tabular}{cc}
\includegraphics[width=0.5\linewidth]{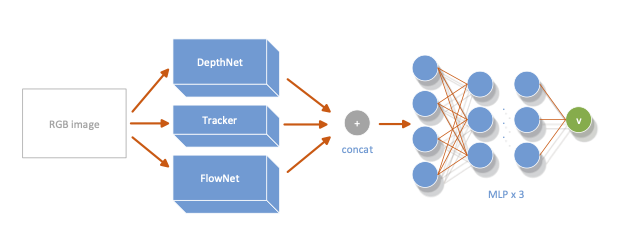} &
\includegraphics[width=0.5\linewidth]{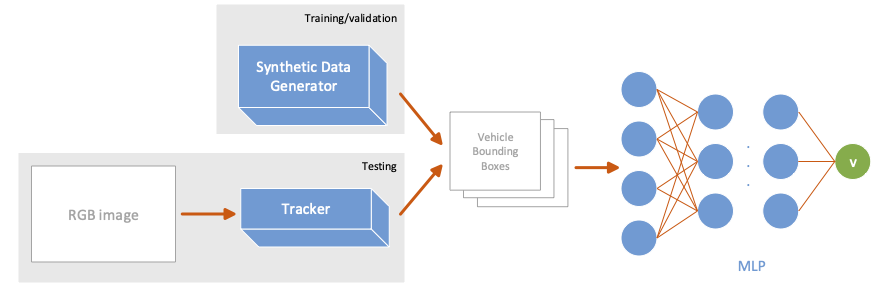}
\end{tabular}
\caption{The state-of-the-art architecture of~\cite{kampelmuhler2018} (top) consists of three different sub-networks and separate models for each of the three distance intervals. Our architecture (bottom) trained on synthetic data is significantly more straightforward, uses a single model and achieves comparable accuracy.}\label{f:model}
\vspace{-10pt}
\end{figure*}

%% file: fig-bb-distribution.tex
\begin{figure*}
\includegraphics[width=\textwidth]{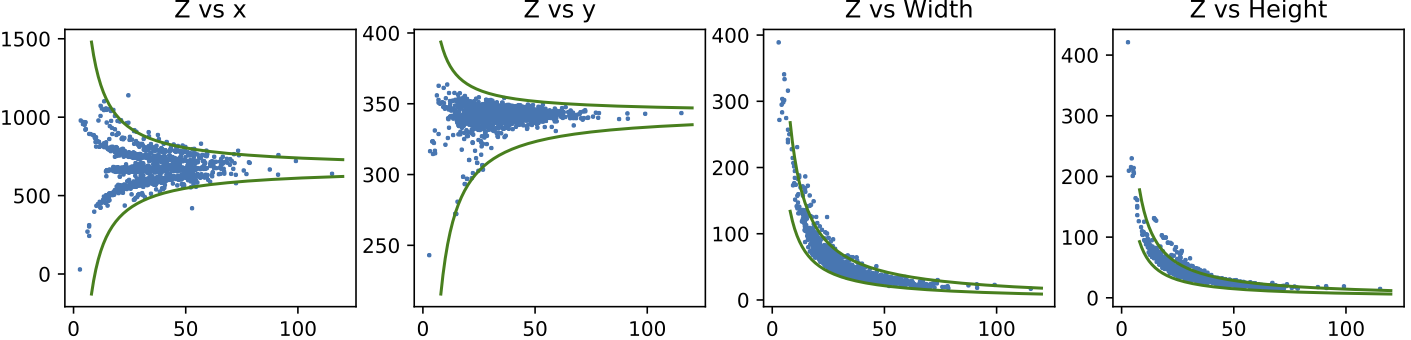}
\vspace{-1.5em}
\caption{%
\textbf{Vehicle statistics.}
Top left: coordinate of the left edge of the bounding box image location vs the depth $Z$ of the enclosed vehicle.
This is constrained by green lines that represent the pixel coordinates of 3D points $9$m left and right of the camera and placed at corresponding depths.
Other plots in order: coordinate $y$ (constraints at $[0,2.5\mathrm{m}]$), box height $h$ ($[1.3\mathrm{m}, 2.5\mathrm{m}]$) and box width $w$ ($[1.5\mathrm{m}, 3\mathrm{m}]$).
}\label{fig:bb_coords}
\end{figure*}

%% file: experiments.tex
\section{Learning from real or synthetic data}\label{sec:dataset}

We begin by discussing a real benchmark dataset sufficient to train our model and then we discuss how we can generate analogous synthetic data effectively in the space of object bounding boxes.

\subsection{Real data: TuSimple}\label{sec:real}
\begin{figure*}
\centering
\includegraphics[width=0.85\textwidth]{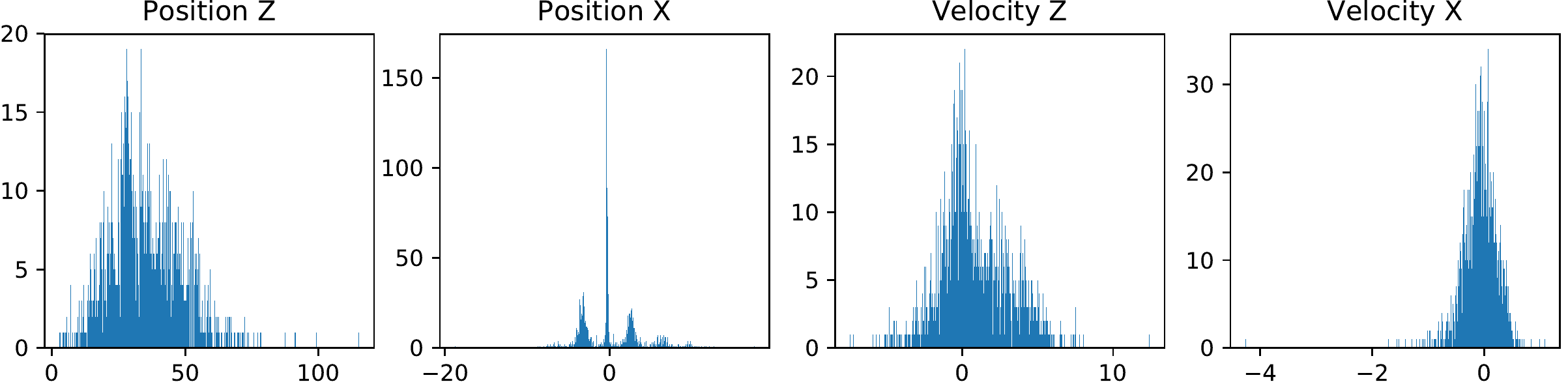}
\vspace{-1em}
\caption{Distribution of vehicle instances in the TuSimple dataset.}%
\label{fig:distributions}
\end{figure*}

While many driving datasets exist in the realm of autonomous driving many focus on single frame tasks such as object detection in 2D/3D, depth estimation, semantic segmentation and localisation. The only dataset with nessecary annotations for our task with comparable work come from TuSimple\cite{CVPR2017Challenge}.
The TuSimple dataset was introduced for the CVPR2017 Autonomous Driving velocity estimation challenge~\cite{CVPR2017Challenge}. It contains $1074$ ($269$) driving sequences with $1442$ ($375$) annotated vehicles in the training (respectively testing) set, split into three subsets based on the distance between the ego-vehicle and the observed car (see~\cref{tbl:datasetsplit}). The dataset was recorded on a motorway by an standard camera (image resolution $1280\times720$) mounted on the roof of the ego-vehicle. Each driving sequence contains 40 frames, captured at 20fps as seen in \cref{f:samples}. LiDAR and radar were simultaneously used to capture the position and velocity of nearby vehicles and recorded data were then used to give the ground truth velocity and 3D position of each vehicle in the last frame of the sequence. Additionally, a manually created bounding box (in the image space) is available in the last frame for each vehicle. Using the notation from~\cref{sec:velocityestimation}, the dataset therefore provides $1442$ training and $375$ testing triplets $(\mathbf{I}, b_T, V)$. As shown in \cref{fig:distributions} the TuSimple dataset spans a wide range of distances from the camera with a high bias towards the same lane as is typical in driving imagery. Velocity in both $Z$ and $X$ is roughly uniformly distributed around $0$ which is expected in motorway situations.

\begin{table}
\centering
\small
\begin{tabular}{l|ccc|c}
\toprule
 & Near & Medium & Far & Total \\
 distance [m] & $d < 20$ & $20 < d < 45$ & $d > 45$ 
 \\
 \midrule
 Train & 166 & 943 & 333 & 1442\\
 Test & 29 & 247 & 99 & 375\\
\bottomrule
\end{tabular}
\caption{The number of annotated vehicles in the TuSimple dataset~\cite{CVPR2017Challenge} and their split based on their distance from the ego-vehicle.}%
\label{tbl:datasetsplit}
\vspace{-2em}
\end{table}

\begin{figure*}[b]
\centering    
\includegraphics[width=0.45\textwidth]{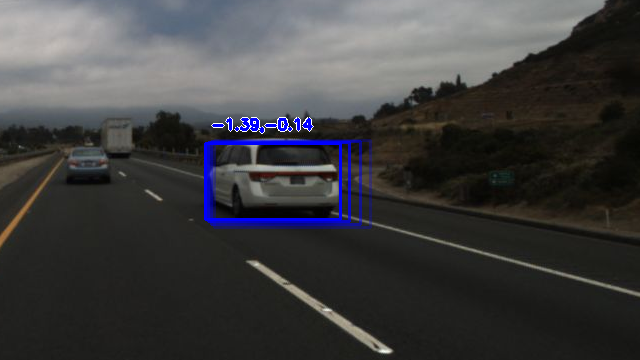}
\includegraphics[width=0.45\textwidth]{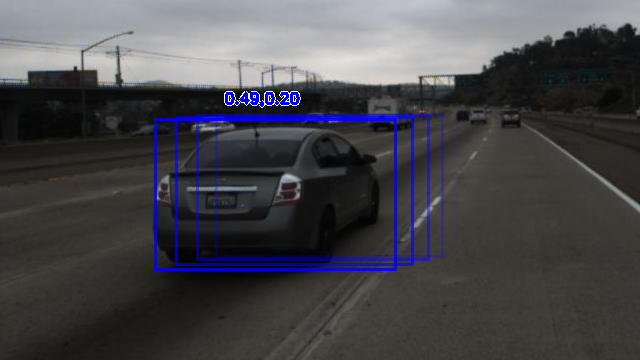}
\caption{Here we show and example of our synthetic boxes in these cases generated starting at the same location as a real object. The fainter boxes depict the location of the object in subsequent frames for the velocity written above which may differ from the actual vehicles path in real data}

\label{fig:data_example}
\end{figure*}

\subsection{Synthetic data}\label{sec:synthetic}

By distilling the information contained in an image to bounding boxes $\mathbf{b}$ we now have data which is highly-interpretable and easy to characterise statistically; the latter can be used to generate \emph{synthetic} training data, which ideally will have very similar statistical properties to the real data. We could also choose to simulate vehicles with speeds not normally seen in the real world to train a network capable of understanding such abnormal situations.

We use the training subset of the TuSimple dataset (see~\cref{sec:dataset}) to infer these statistical properties.
In~\cref{fig:bb_coords}-left we plot the bounding box horizontal coordinates $x$ vs the depth of the enclosed vehicle.
There is some obvious structure in the data; in particular, several lanes to the left and to the right of the ego-vehicle are clearly visible.
The other plots show the $y$\textbf{} coordinate as well as the bounding box width and height.
The latter are highly constrained by the physical sizes of vehicles.

\begin{figure*}
\centering
\includegraphics[width=\textwidth]{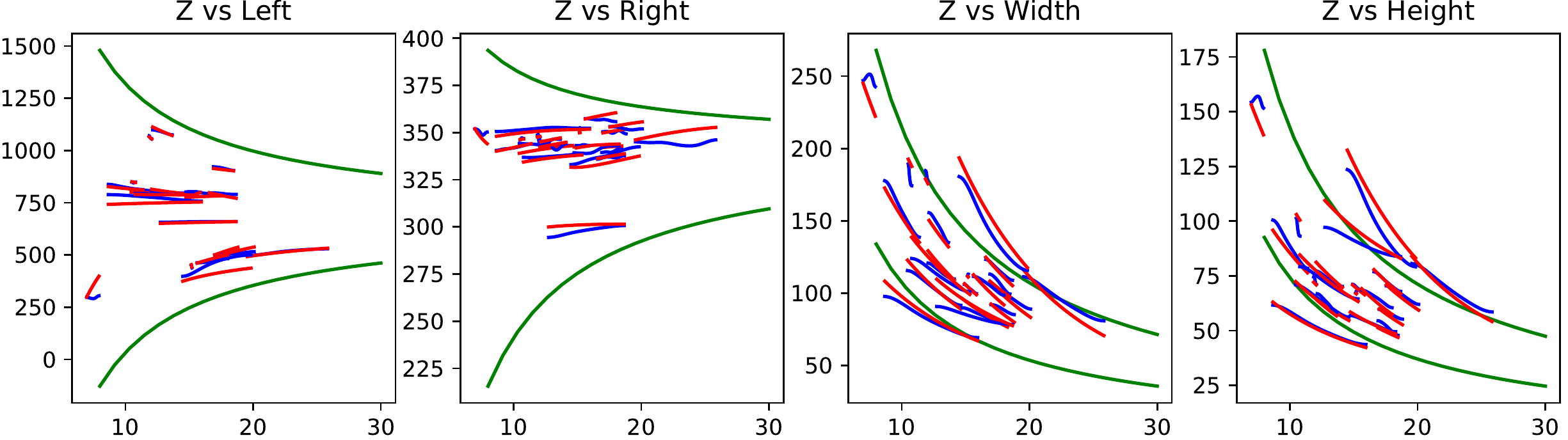}
\caption{\textbf{Real vs synthetic vehicle sequences.} The motion of the bounding box in the video sequence - vehicle motion in real video as captured by the tracker (blue), and synthetic motion generated by our method (red). In the width and height plot we see some tracking inaccuracy when the blue lines increase going to the right.}%
\label{fig:motion}
\end{figure*}

We found learning to be sensitive to the distribution of vehicles locations and less so of car velocities and sizes.
We thus represent the location distribution empirically (\cref{fig:data_example}) and sample from TuSimple the first bounding box in each to obtain its 3D ground point $(X,0,Z)$.
This is then re-projected to the image as explained in~\cref{sec:elementary} to obtain the bounding box location $(x,y)$.
The box height and width are obtained indexing with the depth $Z$ polynomial fits to the TuSimple height/width data (\cref{fig:bb_coords}).
Finally, the track is simulated by sampling a velocity vector $V$ from a Gaussian fit of the TuSimple velocity data (\cref{fig:distributions}) and integrating the motion (\cref{fig:motion}).

Note that only the bounding box positions are empirical, whereas the other parameters are sampled form very simple distributions.
We show that this is sufficient to train models that achieve excellent performance on real data.
This helps isolating what are the important/sensitive priors (location) and what are not (size, velocity) for velocity estimation.


\section{Evaluation}\label{sec:evaluation}

We evaluated our approach using two models.
For both models, we used MLP with 4 hidden layers, trained for 150 epochs, using dropout of 0.2 after the Concatenated ReLU\cite{CReLU} activation function, learning rate with learning rate $6\times10^{-4}$ adjusted using exponential decay with decay rate set to $0.99$.

In the first model denoted as \textit{MLP-tracker}, we initially processed all video frames by the MedianFlow~\cite{MedianFlow} tracker, using the bounding box $b_T$ from the ground truth for tracker initialization. We then used the resulting sequences $\mathbf{b}=(b_1,\dots,b_T)$ as the training data for the MLP (see~\cref{sec:velocityestimation}).
At test time, we followed the same procedure, which is pre-processing the input video sequence by the tracker, and then feeding the intermediate representation $\mathbf{b}$ into the MLP to obtain the velocity estimate.

We follow the TuSimple competition protocol, and report velocity estimation accuracy $E_v$ calculated as the average over the three distance-based subsets (\cref{tbl:datasetsplit}):
\begin{equation}
E_v = \frac{E_{v}^{\mbox{\tiny{near}}} + E_{v}^{\mbox{\tiny{medium}}}+E_{v}^{\mbox{\tiny{far}}}}{3},\qquad
E_v^S = \frac{1}{|S|}\sum_{i \in S} = \|V_i - \hat{V}_i\|^2,
\end{equation}
where $V_i$ is the ground truth velocity from the LiDAR sensor and $\hat{V}_i$ is method's velocity estimate. 
We also note that according to the dataset authors, the overall ground-truth accuracy is at around 0.71m/s, however the accuracy will almost certainly depend on the distance of the observed car.

Using the above model, we reach the overall velocity estimation error of $1.29$, which  outperforms the state-of-the-art\footnote{Kampelm\"uhler~\etal also report the error of $1.86$ and $1.25$ for their method using only tracking information, however this was not the competition entry and it is not obvious from the paper how the latter number was reached and what data were used for training.} and significantly more complex method of Kampelm\"uhler~\etal\cite{kampelmuhler2018}. This experiment also shows, that the intermediate representation $\mathbf{b}$ contains sufficient amount of information to successfully infer vehicle velocity.

In the second model \textit{MLP-synthetic}, we instead used the synthetic training data  as the intermediate representation $\mathbf{b}$. We generated $11536$ samples using the generation procedure described in \cref{sec:synthetic} and used them as the training samples for the MLP. At testing time, we again used the MedianFlow tracker to get the intermediate representation $\mathbf{b}$ from real video sequences and fed them into the MLP to obtain vehicle velocity estimates (see \cref{f:model} bottom). Because our synthetic boxes are generated without noise inherently coming from the tracker road surface and other sources, at test time we apply a Gaussian filter with $\sigma=5$ in each coordinate of the bounding box as pre-processing to remove noise.

The resulting model has the velocity estimation error of $1.28$, with the biggest improvement in the Far range (see \cref{tbl:results}). Generating more synthetic data for training did not result in further improvement in accuracy, which we contribute to the relatively small size of the testing set and the relative low variance of vehicle speed owing to motorway driving style.





\begin{table}[h]
\centering
\small
\begin{tabular}{l|l|lll}
\toprule
 & $E_v$  & $E_{v}^{\mbox{\tiny{near}}}$ & $E_{v}^{\mbox{\tiny{medium}}}$  & $E_{v}^{\mbox{\tiny{far}}}$ \\
 \midrule
Kampelm\"uhler~\etal\cite{kampelmuhler2018} (1st) & 1.30  &    0.18    &   \textbf{0.66}  &     3.07      \\
Wrona (2nd) & 1.50 & 0.25 & 0.75 & 3.5 \\
Liu (3rd) & 2.90 & 0.55 & 2.21 & 5.94 \\
\midrule
geometric reprojection & 8.5 & 0.48 & 1.50 & 23.60\\ 
Ours (MLP-tracker) & 1.29 & 0.18 & 0.70 & 2.99 \\
Ours (MLP-synthetic) & \textbf{1.28} & \textbf{0.17} & 0.72 & \textbf{2.96}\\ 
\midrule
\textit{LiDAR} & \multicolumn{4}{c}{\textit{0.71}} \\
\bottomrule
\end{tabular}
\caption{Vehicle velocity estimation error $E_v$ on the TuSimple dataset, including the top 3 best-performing methods of the CVPR 2017 Vehicle Velocity Estimation challenge~\cite{CVPR2017Challenge}. The average accuracy of the LiDAR sensor used for ground truth acquisition added for reference. Runtime of Kampelm\"uhler~\etal\cite{kampelmuhler2018} is $423ms$ owing to the varying and complex inputs required for a forward pass, our method can perform a forward pass in $10ms$ on the same hardware(runtime for other methods are not public).}\label{tbl:results}
\end{table}

%% file: conclusion.tex
\section{Conclusion}\label{sec:conclusion}
In this paper, we showed how to efficiently predict the velocity of other vehicles based only on a video from a standard monocular camera by introducing an intermediate representation which decouples perception from velocity estimation. We also showed how priors in real data can be exploited to generate synthetic data in the intermediate representation and that such synthetic training data can be used to build a system capable of processing real-world data without any loss in accuracy.

The decomposition is advantageous not only in terms of the ability to easily generate training data with the desired parameters, but also to model different driving styles or various emergency situations. Last but not least, it also ensures each component is individually verifiable, which is essential to meet reliability standards for autonomous driving.